\documentclass{article}
\usepackage{ijcai26}

\usepackage{times}
\usepackage{soul}
\usepackage{url}
\usepackage[hidelinks]{hyperref}
\usepackage[utf8]{inputenc}
\usepackage[small]{caption}
\usepackage{natbib}
\usepackage{graphicx}
\usepackage{amsmath}
\usepackage{amsthm}
\usepackage{multirow}
\usepackage{booktabs}
\usepackage{algorithm}
\usepackage{algorithmic}
\usepackage[switch]{lineno}
\usepackage{amssymb}
\usepackage{CJKutf8}
\usepackage{color}
\title{SPIRIT: Adapting Vision Foundation Models for Unified Single- and Multi-Frame Infrared Small Target Detection}

\author{
Qian Xu$^1$
\and
Xi Li$^1$\and
Fei Gao$^{1}$\and
Jie Guo$^1$\and
Haojuan Yuan$^4$\and 
Shuaipeng Fan$^1$\And
Mingjin Zhang$^1$
\affiliations
$^1$Xidian University~~~~~~~~~$^2$Shanghai Academy of Spaceflight Technology\\
}
\begin{document}
\begin{CJK}{UTF8}{gkai}
\maketitle

\begin{abstract}
% Infrared small target detection (IRSTD) is a core capability for long-range surveillance, yet remains challenging due to the physics-dominated imaging process: targets occupy only a few pixels with weak contrast, while backgrounds are highly structured and dynamic. Recently, vision foundation models (VFMs) pretrained on large-scale RGB data offer strong transferable representations, but directly adapting semantics-oriented VFMs to IRSTD often fails in two ways: (i) tiny target cues are submerged by deep feature aggregation, and (ii) appearance-only temporal association in videos is under-constrained, leading to spurious matches and amplified false alarms.
% We propose {PST-Former}, a unified physics-informed spatiotemporal framework that adapts a hierarchical VFM backbone to both single-frame and video IRSTD without modifying backbone blocks. Spatially, we introduce a Physics-Informed Feature Refinement (PIFR) that performs feature-domain background--target disentanglement via an efficient rank-1 projection and learnable group shrinkage, enhancing sparse target  under clutter. Temporally, we design PGMA, a prior-guided memory attention mechanism that biases memory cross-attention with a differentiable feasibility field induced by previous-frame detections, enabling feasibility-constrained  retrieval under infrared appearance ambiguity. Extensive experiments on multiple IRSTD benchmarks demonstrate that PST-Former consistently improves VFM-based baselines and achieves  state-of-the-art performance.

Infrared small target detection (IRSTD) is crucial for surveillance and early-warning, with deployments spanning both single-frame analysis and video-mode tracking. A practical solution should leverage vision foundation models (VFMs) to mitigate infrared data scarcity, while adopting a memory-attention-based temporal propagation framework that unifies single- and multi-frame inference. However, infrared small targets exhibit weak radiometric signals and limited semantic cues, which differ markedly from visible-spectrum imagery. This modality gap makes direct use of semantics-oriented VFMs and appearance-driven cross-frame association unreliable for IRSTD: hierarchical feature aggregation can submerge localized target peaks, and appearance-only memory attention becomes ambiguous, leading to spurious clutter associations.
To address these challenges, we propose SPIRIT, a unified and VFM-compatible framework that adapts VFMs to IRSTD via lightweight physics-informed plug-ins. Spatially, PIFR refines features by approximating rank-sparsity decomposition  to suppress structured background components and enhance sparse target-like signals. Temporally, PGMA injects history-derived soft spatial priors into memory cross-attention to constrain cross-frame association, enabling robust video detection while naturally reverting to single-frame inference when temporal context is absent. Experiments on multiple IRSTD benchmarks show consistent gains over VFM-based baselines and SOTA performance.
\end{abstract}

\section{Introduction}

Infrared small target detection (IRSTD) is a fundamental component of long-range surveillance and early-warning systems, with applications in airborne defense, maritime monitoring, and space situational awareness. In realistic deployments, systems may switch between continuous tracking (video mode) and discrete situational analysis (single-frame mode). This dual requirement motivates a unified detection architecture that works effectively in both regimes and can gracefully degrade to single-frame inference when temporal information is unavailable. Unlike generic object detection, IRSTD operates in a radiometry-driven regime where targets are tiny and nearly textureless, yet backgrounds are complex and highly structured. As a result, detection depends mostly on subtle intensity contrast; in the video setting, modest but informative motion cues further help, while color, texture, and shape cues are largely absent.  Traditional model-driven methods rely on handcrafted priors, such as local contrast or low rank sparse decomposition, but are prone to failure in the presence of strong clutter. Data-driven deep learning methods based on CNNs or Transformers, on the other hand, are often limited by the scarcity of high-quality labeled infrared data. Moreover, most infrared video detectors rely on fixed-clip spatiotemporal models, which are rigid and cannot naturally support both video and single-frame inference within one model. By contrast, memory-attention-based temporal propagation provides a natural route to a unified design: it exploits history when available for video inference, yet naturally reverts to single-frame inference when temporal context is absent. Meanwhile, IRSTD also demands strong generic representations under limited infrared annotations, motivating the use of large-scale pretrained models.
\begin{figure}[t]
  \centering
  \includegraphics[width=1\linewidth]{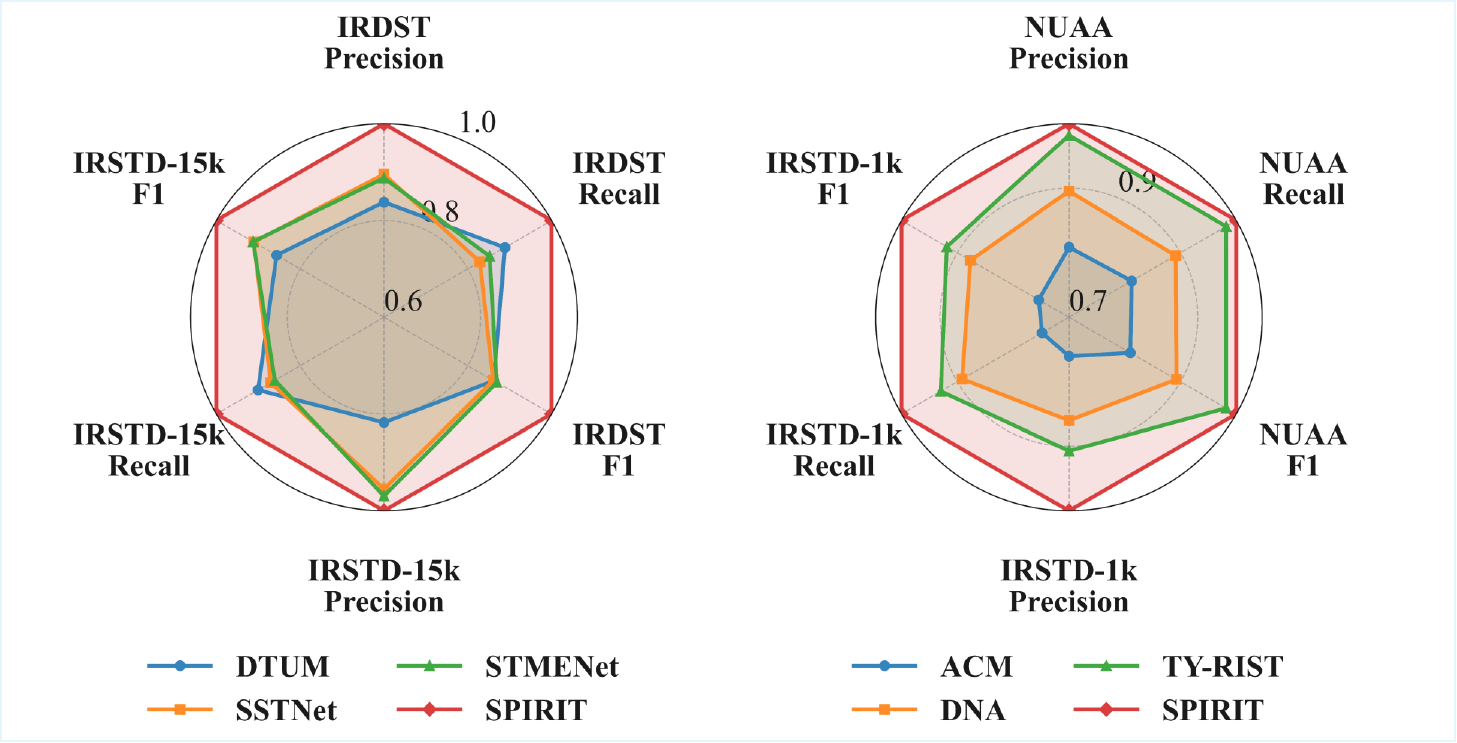}
  \caption{Performance comparison on multi-frame (left) and single-frame (right) IRSTD datasets. All metrics are normalized for clear visualization. Larger enclosed areas indicate superior performance.}
  \label{fig:rader}
\end{figure}

Advances in vision foundation models (VFMs), exemplified by SAM~\cite{ravi2024sam} and DINO-style architectures~\cite{simeoni2025dinov3}, are redefining the  visual recognition pipeline. Instead of training task-specific backbones from scratch, modern detectors and segmenters increasingly adapt a single large-scale model pretrained on web-scale RGB images. These VFMs provide strong generic features and have demonstrated remarkable transferability across a wide range of downstream tasks, particularly those operating on natural images with rich semantic content. In principle, such models could mitigate data scarcity in IRSTD and offer stronger contextual representations, making the use of VFMs as backbones for IRSTD an attractive strategy.

However, directly transplanting these semantics-oriented VFMs to the physics-dominated IRSTD domain reveals a fundamental modality mismatch. We identify two critical failure modes where the inductive biases of VFMs conflict with the nature of infrared imaging:
\textbf{(i) Spatial feature submergence.} Hierarchical backbones progressively aggregate information through deep token mixing and large effective receptive fields. While this benefits semantic coherence in natural images, IR small targets manifest as weak, localized intensity singularities occupying only a few pixels, often embedded in structured clutter such as cloud edges or sea waves. Under repeated aggregation, such small peaks are easily diluted and overshadowed by strong background responses. Consequently, the signal-to-clutter ratio (SCR) of valid targets collapses in deeper layers, leading to missed detections.
\textbf{(ii) Temporal appearance ambiguity.} Along the temporal dimension, Transformers and memory-attention-based architectures typically perform association by matching appearance features across frames, which is effective when objects have distinctive semantic attributes. However, in infrared sequences, small targets are nearly textureless. Their appearance is  inherently ambiguous and often indistinguishable from dynamic clutter.  A global, appearance-only attention mechanism is therefore under-constrained and cannot reliably query the correct target from memory,  forming spurious associations between high-response background patterns and the target trajectory that amplify false alarms and degrade detection stability.
In short, semantics-driven representations tend to dilute weak local peaks spatially and become unreliable for appearance-based association temporally.

To address these challenges, we propose SPIRIT ({S}patio-temporal {P}hysics-{I}nformed {R}epresentation for {I}nfrared {T}argets), a unified framework that adapts semantics-oriented VFMs to IRSTD by injecting physically motivated constraints into feature representation and attention, in a backbone-compatible manner and without redesigning the pretrained VFM. The proposed modules are lightweight plug-ins that can be integrated into different VFM backbones with minimal changes. On the spatial side, we introduce a Physics-Informed Feature Refinement (PIFR) that acts as a differentiable physical filter inside the encoder. PIFR approximates rank-sparsity decomposition in the feature domain, suppressing low-rank background components and enhancing sparse target-like responses, thereby alleviating feature submergence without sacrificing the benefits of large-scale pretraining. On the temporal side, we design a Prior-Guided Memory Attention (PGMA) mechanism that injects a feasibility prior into memory cross-attention: cues distilled from previous-frame detections are encoded as a soft spatial prior and used to bias the attention toward plausible associations.This converts unconstrained global appearance matching into feasibility-guided local matching, suppressing clutter that is visually similar but spatially implausible. SPIRIT operates in both single-frame and video regimes, and when temporal information is unavailable it naturally falls back to a purely spatial detector, enabling flexible deployment under dynamic operational requirements.

Our main contributions are summarized as follows:
\begin{itemize}
% \item We propose {PST-Former}, a unified physics-informed adaptation framework that transfers a pretrained hierarchical  vision foundation backbone to both single-frame and video IRSTD, without modifying the backbone blocks.
% \item We introduce {PIFRM}, a plug-and-play feature refinement module that performs differentiable background--target disentanglement via a low-rank--plus--sparse relaxation in the feature domain, enhancing weak small-target evidence under structured clutter.
% \item We design {PGMA}, a prior-guided memory attention mechanism that biases memory cross-attention using a differentiable feasibility field induced by previous-frame detections, and further employs confidence-gated memory writing to prevent error reinforcement in infrared videos.
\item We propose SPIRIT, a unified and VFM-compatible framework that injects physics-informed plug-ins into semantics-oriented VFMs, supports both single-frame and video IRSTD, and gracefully degrades to single-frame inference when temporal cues are unavailable.

\item We introduce PIFR, a differentiable rank-sparsity  decomposition refinement module that suppresses structured backgrounds and enhances sparse targets to mitigate feature submergence.

\item We design PGMA, a prior-guided memory attention mechanism that leverages history-derived spatial priors to constrain cross-frame association, reducing spurious matches and improving detection stability.
\item Extensive experiments on  IRSTD benchmarks demonstrate that SPIRIT consistently improves VFM-based baselines and achieves SOTA performance.

\end{itemize}

\section{Related Work}
\paragraph{Single-frame IR Small Target Detection.}
Early IRSTD mainly relied on model-driven priors such as local contrast and low-rank/sparse decomposition~\cite{gao2013infrared,zhang2018infrared,dai2017reweighted}, which are interpretable but fragile under strong structured clutter.
Recent CNN/Transformer detectors~\cite{dai2021asymmetric,li2022dense,zhang2025irmamba} learn features end-to-end and achieve notable gains, yet they are typically trained from scratch on scarce infrared annotations, limiting generalization.
This motivates transferring pretrained VFMs to IRSTD.

\paragraph{Video IRSTD and Temporal Association.}
Temporal cues help suppress clutter via motion consistency; classical pipelines use flow compensation or temporal filtering.
Deep video IRSTD models~\cite{yan2023stdmanet,tong2024st,chen2024sstnet} aggregate multi-frame features with 3D convolutions or spatiotemporal attention, but many operate on fixed-length clips and cannot gracefully degrade to single-frame inference.
Moreover, generic video detectors often associate frames by appearance similarity, which is under-constrained for textureless infrared targets and easily confused by dynamic clutter, motivating feasibility-guided temporal matching beyond appearance-only attention.
\begin{figure*}[t]
  \centering
  \includegraphics[width=0.75\linewidth]{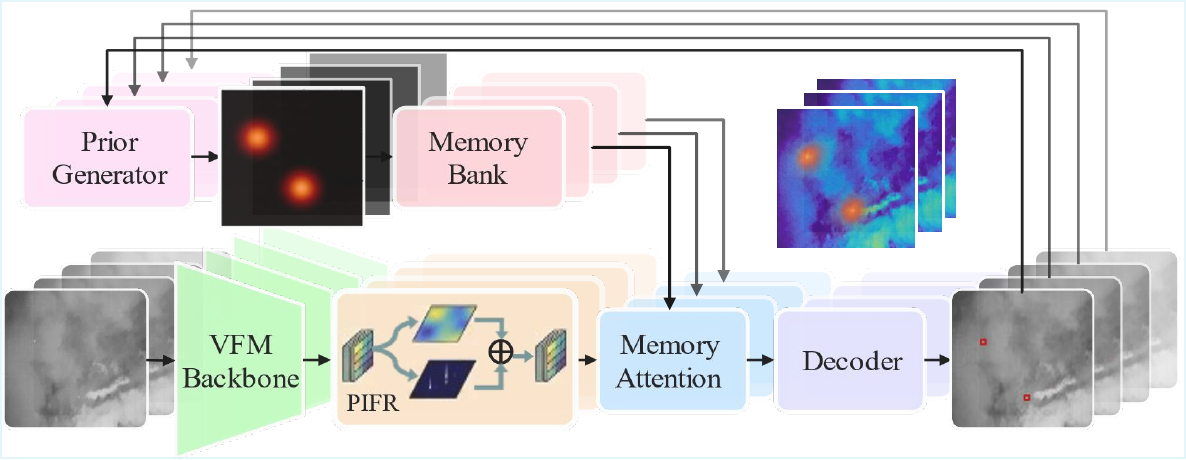}
  \caption{The overall architecture of the proposed method. It integrates PIFR into the VFM backbone to decouple background and target for feature enhancement, and employs prior guided memory attention to suppress spurious associations by leveraging a feasibility field to focus on physically plausible regions. }
  \label{fig:architecture}
\end{figure*}
% \subsection{Vision Foundation Models and Adaptation}
% \label{sec:related_work_vfm}
\paragraph{Vision Foundation Models and Adaptation.} Vision foundation models (VFMs) such as SAM~\cite{ravi2024sam} and DINO-style models~\cite{simeoni2025dinov3} provide transferable representations widely used for downstream vision tasks.
Parameter-efficient fine-tuning methods including LoRA, adapters, and prompt tuning enable efficient adaptation.
Recent works have explored transferring VFMs to infrared small-target tasks~\cite{zhang2024irsam,zhang2025saist}, but also report that direct transfer can be unstable in cluttered scenes with limited gains due to the modality gap.
Thus, effective VFM adaptation for IRSTD likely requires task-aware designs beyond routine fine-tuning.

\section{Method}
\label{sec:method}

\paragraph{Overview.}
The overall architecture of SPIRIT is illustrated in Figure~\ref{fig:architecture}. It is built upon a hierarchical VFM backbone to extract multi-scale feature pyramids. To mitigate feature submergence, which typically intensifies during deep feature aggregation, we strategically insert the {PIFR} module into the {last two} backbone stages. This placement allows the model to explicitly restore sparse target signals in high-level semantic features before they are overwhelmed by the background. For video inputs, these refined features serve as queries for the {PGMA} mechanism to suppress spurious associations via history-constrained interaction. Finally, to translate the refined representations into detections, we adopt the lightweight decoder from DEIM~\cite{huang2025deim}. We preserve its core design and only modify the interface to align VFM tokens, ensuring a fair comparison while inheriting its high inference efficiency.

\subsection{Physics-Informed Feature Refinement}
\label{sec:pifr}

\begin{figure}[t]
  \centering
  \includegraphics[width=1\linewidth]{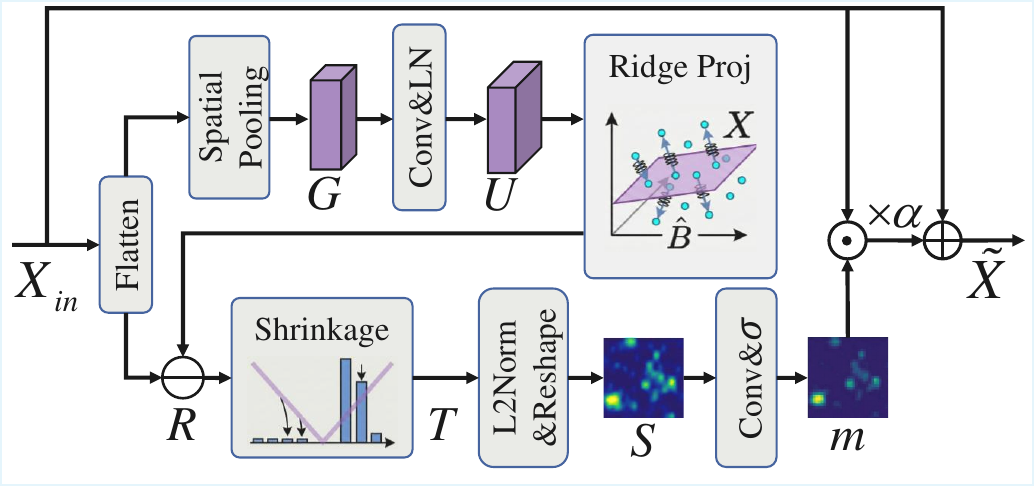}
  \caption{Illustration of the PIFR module, which utilizes ridge projection and shrinkage for feature refinement.}
  \label{fig:pifr}
\end{figure}

% Infrared small targets typically manifest as localized, sparse response spikes with weak semantics, whereas structured background/clutter becomes increasingly globally coherent after hierarchical aggregation in a VFM backbone. This discrepancy often causes feature submergence: target peaks are diluted by token mixing and large effective receptive fields, and are eventually overwhelmed by background responses. To explicitly restore radiometry-driven separability in the feature domain, we introduce Physics-Informed Feature Refinement (as illustrated in Figure~\ref{fig:pifr}), a lightweight plug-in that approximates a low-rank sparse separation and injects the recovered sparse cues back into the backbone in an identity-preserving manner.
Infrared small targets appear as weak sparse spikes, whereas backgrounds become globally coherent after VFM aggregation. This discrepancy causes feature submergence, where target peaks are overwhelmed by deep token mixing. To explicitly disentangle sparse targets from coherent backgrounds, we introduce Physics-Informed Feature Refinement (Figure~\ref{fig:pifr}). This lightweight plug-in approximates a rank-sparsity decomposition to recover target cues and injects them back via an identity-preserving residual.

Let $X_{\text{in}}\in\mathbb{R}^{C\times H\times W}$ be a deep feature map and $\bar{X}\in\mathbb{R}^{C\times N}$ ($N=HW$) its spatially flattened form. Following a standard infrared prior, we model
\begin{equation}
\bar{X}=B+T+N,
\label{eq:pifr_decomp}
\end{equation}
where $B$ denotes structured background/clutter, $T$ denotes sparse target-like responses, and $N$ denotes residual noise. Classical RPCA estimates $B$ via iterative SVD, which is costly and inconvenient to integrate into modern VFM backbones. Instead, we exploit an empirical spectral property of infrared backgrounds: after deep aggregation, clutter responses are dominated by a few globally correlated components, yielding a rapidly decaying singular-value spectrum \cite{gao2013infrared,zhang2018infrared}. We thus approximate the background by a fixed low-rank subspace of rank $r$ (we use $r=4$ throughout, as most background energy concentrates in the leading singular directions),
\begin{equation}
B \approx U V^\top,\enspace \mathrm{rank}(B)\le r.
\label{eq:pifr_rankr}
\end{equation}

To avoid explicit SVD, we construct coarse background prototypes via spatial pooling. Let $M\in\mathbb{R}^{N\times r}$ be a fixed pooling matrix whose columns correspond to $r$ non-overlapping coarse spatial bins, normalized to average features within each bin. We compute pooled prototypes
\begin{equation}
G=\bar{X}M\in\mathbb{R}^{C\times r},
\label{eq:pifr_proto}
\end{equation}
map $G$ to a background basis $U=\phi(G)\in\mathbb{R}^{C\times r}$ using a lightweight projector $\phi(\cdot)$, and estimate the low-rank background by a closed-form ridge projection,
\begin{equation}
\hat{B}=U\,(U^\top U+\delta I)^{-1}U^\top \bar{X},
\label{eq:pifr_proj}
\end{equation}
where $\delta$ is a small constant for numerical stability and $I$ is the $r\times r$ identity. This operation is SVD-free and lightweight, requiring only an $r\times r$ matrix inverse. We then form the residual $R=\bar{X}-\hat{B}$, which contains sparse/high-frequency components where small targets are expected to reside. To extract target-like responses while suppressing diffuse noise, we apply a token-wise group shrinkage,
\begin{equation}
T_{:,i}=\max\!\Big(1-\frac{\mathrm{softplus}(\rho)}{\|R_{:,i}\|_2+\epsilon},0\Big)\,R_{:,i},\enspace i=1,\dots,N,
\label{eq:pifr_shrink}
\end{equation}
where $\rho$ is learnable and $\epsilon$ is a small constant. Using $\mathrm{softplus}(\rho)$ ensures a positive shrinkage threshold and stabilizes optimization: locations with small $\ell_2$ magnitude across channels are attenuated, while sparse peaks with large magnitude are preserved. 
% To convert the recovered sparsity into a spatial modulation signal, we compute the token energy map
We then convert sparsity into a soft gate by
\begin{equation}
S=\mathrm{reshape}\big(\{\|T_{:,i}\|_2\}_{i=1}^{N}\big),\enspace
m=\sigma(\mathrm{Conv}(S)),
\label{eq:pifrm_gate}
\end{equation}
and refine features in an identity-preserving residual form:
\begin{equation}
\tilde{X}=X_{\text{in}}+\alpha\,(X_{\text{in}}\odot m),
\label{eq:pifrm_inject}
\end{equation}
where $\odot$ denotes element-wise multiplication and $\alpha$ is a learnable scale initialized to $0$. This initialization preserves the pretrained backbone at early training stages while gradually introducing the physical prior, improving stability when adapting large VFMs to the infrared domain. Overall, PIFR counteracts feature submergence by suppressing globally coherent clutter and enhancing sparse target evidence, while introducing minimal overhead dominated, making it practical for integration into diverse hierarchical VFM backbones.

\begin{table*}[t] 
  \centering
 
  \begin{tabular}{l c cccc cccc r r r} % +1列FPS
    \toprule
    \multirow{2}{*}{Method}
    &\multirow{2}{*}{Venue}
    & \multicolumn{4}{c}{\textbf{IRDST}}
    & \multicolumn{4}{c}{\textbf{IRSTD-15k}} 
    & \multirow{2}{*}{Params} & \multirow{2}{*}{FLOPs} & \multirow{2}{*}{FPS} \\ % 增加FPS表头
    \cmidrule(lr){3-6} \cmidrule(lr){7-10} 
    & & AP50 & P & R & F1 & AP50 & P & R & F1 &  &  &  \\ 
    \midrule
    % DNANet          & TIP'2022  & 63.61 & 82.92 & 77.48 & 80.11 & 70.46 & 88.55 & 80.73 & 84.46 & 7.2  & 135.2 & - \\
    % ISNet           & CVPR'2022 & 59.78 & 80.24 & 75.08 & 77.58 & 62.29 & 83.46 & 75.32 & 79.18 & 3.5  & 256.7 & - \\
    % UIUNet          & TIP'2022  & 56.38 & 80.95 & 70.29 & 75.25 & 65.15 & 84.07 & 78.39 & 81.13 & 53.1 & 456.7 & - \\
    % AGPCNet         & TAES'2023 & 59.21 & 79.47 & 75.51 & 77.44 & 67.27 & 91.19 & 74.77 & 82.16 & 14.9 & 366.2 & - \\
    % MSHNet          & CVPR'2024 & 63.21 & 82.31 & 77.64 & 79.91 & 60.82 & 89.69 & 68.44 & 77.64 & 6.6  & 69.6  & - \\
    % % SCTransNet      & TGRS'2024 & 92.35 & 96.80 & 90.18 & 93.37 & 63.61 & 86.24 & 74.29 & 79.82 & -    & -     & - \\
    % PConv           & AAAI'2025 & 91.88 & 96.21 & 95.91 & 96.06 & 61.19 & 88.93 & 69.69 & 78.14 & 23.8 & 88.6  & - \\
    % \midrule
    DTUM            & TNNLS'2023& 71.48 & 82.87 & 87.79 & 85.26 & 67.97 & 77.95 & 88.28 & 82.79 & \textbf{9.6}  & 128.2 & 13.9 \\
    TMP             & ESA'2024  & 70.01 & 86.65 & 81.36 & 83.92 & 77.73 & 92.97 & 84.74 & 88.67 & 16.4 & {92.9}  & 25.0 \\
    SSTNet          & TGRS'2024 & 71.55 & 88.56 & 81.92 & 85.11 & 76.96 & 91.05 & 85.29 & 88.07 & 11.9 & 123.6 & 22.6\\
    Tridos          & TGRS'2024 & 73.72 & 84.49 & 89.35 & 86.85 & 80.41 & 90.71 & 90.60 & 90.65 & 14.1 & 130.7 & 13.7 \\
    STMENet         & ESA'2025  & 73.40 & 87.78 & 84.22 & 85.96 & 77.33 & 92.42 & 84.35 & 88.21 & \underline{10.4} & \textbf{45.8}  & \underline{48.5} \\
    MOCID           & AAAI'2025 & 94.74 & \textbf{98.92} & 96.86 & 97.88 & 86.84 & 90.37 & 92.44 & 91.40 & 13.1 & 98.7  & 11.5 \\
    TDCNet          & AAAI'2025 & \underline{94.79} & 98.12 & \underline{97.71} & \underline{97.91} & \underline{88.17} & \underline{94.50} & \underline{96.75} & \underline{95.61} & 24.8 & 95.7  & 18.5 \\
    SPIRIT           & --        & \textbf{96.24} & \underline{98.91} & \textbf{98.26} & \textbf{98.58} & \textbf{89.56} & \textbf{95.34} & \textbf{97.15} & \textbf{96.23} & 32.2    & \underline{65.6}     & \textbf{53.1} \\
    \bottomrule
  \end{tabular}
  
   \caption{Comparison on multi-frame benchmarks (IRDST and IRSTD-15k). The best results are highlighted in bold and the second-place results are highlighted in underline.}
   \label{tab:multi-frame}
\end{table*}

\subsection{Prior-Guided Memory Attention}
\label{sec:pgma}

Infrared video small targets are extremely small and often textureless; consequently, appearance similarity alone is unreliable for temporal association. Meanwhile, dynamic clutter can produce target-like responses across frames, so unconstrained memory retrieval based purely on global appearance matching may form spurious associations. Worse, such mistakes can be progressively reinforced through the recurrent write--read cycle of a memory bank, leading to temporal drift and persistent false alarms. To mitigate this IR-specific failure mode, we propose Prior-Guided Memory Attention (PGMA), which formulates temporal association as feasibility-constrained memory retrieval with a prior-gated memory encoding. When the memory is empty or the feasibility prior degenerates to uniform, PGMA naturally reduces to standard cross-attention, unifying single-frame and video inference without additional branches.

The refined deepest feature $\tilde{X}_t^{(L)}\in\mathbb{R}^{C\times H_L\times W_L}$ is adopted as the carrier for temporal interaction, flattened into $N=H_LW_L$ tokens to serve as current-frame queries. To construct a continuous and differentiable feasibility prior, previous-frame detections $\mathcal{D}_{t-1}=\{(b_{t-1}^{(m)}, s_{t-1}^{(m)})\}_{m=1}^{M_{t-1}}$ are utilized, where $b^{(m)}=(x_1,y_1,x_2,y_2)$. Each box is mapped to the deep grid to yield its center $c^{(m)}$ and scale $r^{(m)}$, from which a multi-peak feasibility field is generated:
\begin{equation}
G_{t-1}(p)
= \sum_{m=1}^{M_{t-1}} \exp\!\left(-\frac{\|p-c^{(m)}\|_2^2}{2\kappa^2 (r^{(m)})^2}\right),
\end{equation}
where $p$ denotes a grid location and $\kappa$ controls the tolerated search radius. This representation naturally supports an unknown number of targets via a soft union of peaks. When no reliable detection exists ($M_{t-1}=0$), we set $G_{t-1}(p)\equiv 1$, so the prior degenerates to uniform and does not block newly appearing targets.

$G_{t-1}$ is downsampled to match the resolution of $\tilde{X}_{t-1}^{(L)}$, yielding a lightweight gate map:
\begin{equation}
g_{t-1}=\sigma\!\big(\mathrm{Conv}_{3\times 3}(\mathrm{Down}(G_{t-1}))\big)\in[0,1]^{1\times H_L\times W_L},
\label{eq:pgma_gate_map}
\end{equation}
which modulates the memory feature through a gated residual encoding:
\begin{equation}
Z_{t-1}=\tilde{X}_{t-1}^{(L)}+\beta\big(\tilde{X}_{t-1}^{(L)}\odot g_{t-1}\big),
\label{eq:pgma_mem_encode}
\end{equation}
where $\odot$ is element-wise multiplication and $\beta$ is a learnable scale initialized to $0$ to inherit pretrained representations stably. $Z_{t-1}$ is written into  the memory bank $\mathcal{M}_{t-1}$, and the grid coordinate $P_i$ is recorded for each spatial location to enable prior evaluation during addressing. The memory bank $\mathcal{M}_{t-1}$ stores spatial memories from the most recent $K$ frames in a FIFO manner. 

\begin{figure}[t]
  \centering
  \includegraphics[width=1\linewidth]{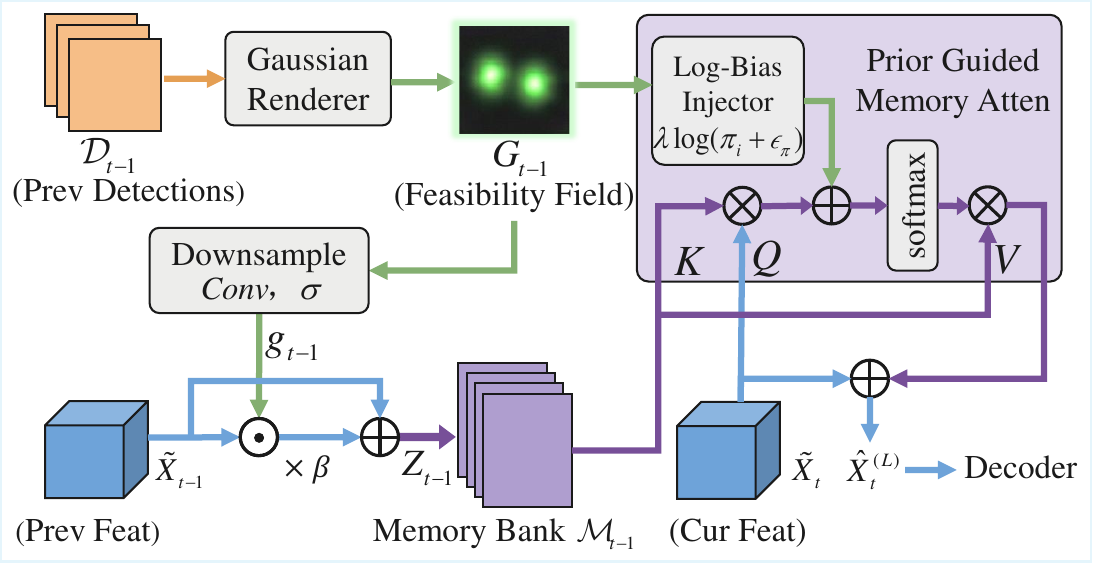}
  \caption{The workflow of PGMA. By utilizing a Gaussian feasibility field as a prior for gated encoding and bias injection, this mechanism guides the model to focus on physically plausible regions to suppress spurious associations. }
  \label{fig:Qualitative}
\end{figure}

At read time, PGMA performs cross-attention from the memory bank: queries $q_j$ come from the current frame tokens, and keys/values $(k_i,v_i)$ come from memory entries. We take appearance similarity as the base score and introduce the feasibility prior as a log-bias. Let $\pi_i = G_{t-1}(P_i)$ be the feasibility weight at the memory position $P_i$. The prior-guided logits and attention weights are:
\begin{equation}
s'_{j,i}=\frac{q_j^\top k_i}{\sqrt{d}}+\lambda\log(\pi_i+\epsilon_\pi),\enspace
w_{j,i}=\mathrm{softmax}_i(s'_{j,i}),
\label{eq:pgma_bias}
\end{equation}
where $\epsilon_\pi$ is a numerical stabilizer. Rather than tuning hyperparameters, we learn the prior strength via a bounded parameterization:
\begin{equation}
\lambda=\lambda_{\max}\sigma(\theta_\lambda).
\label{eq:pgma_lambda_bound}
\end{equation}
For training stability, we initialize $\theta_\lambda$ to a negative value so that $\lambda$ start small. When $\pi_i$ is small, the bias suppresses attention to physically implausible locations, reducing spurious associations caused by appearance ambiguity. Crucially, this soft-constraint mechanism also prevents error accumulation from false alarms (FP): even if an FP induces a misleading spatial bias, the low appearance similarity of suppressed background features (via PIFR) counteracts the bias, preventing the solidification of a false trajectory. Finally, we read memory values and fuse them back to current tokens in a residual form:
\begin{equation}
\hat{q}_j=q_j+\gamma\sum_i w_{j,i}v_i,\enspace
\hat{X}_t^{(L)}=\mathrm{reshape}(\{\hat{q}_j\}_{j=1}^{N}),
\label{eq:pgma_fuse}
\end{equation}
where $\gamma$ is a learnable scale initialized to a small value. The decoder always receives the memory-conditioned current-frame feature $\hat{X}_t^{(L)}$. This architecture ensures robustness to missed detections (FN): when the memory is empty or an FN occurs ($G_{t-1}(p)\equiv 1$), PGMA naturally reverts to global content-based retrieval. This allows the model to recover targets from longer-term history based on appearance cues, ensuring graceful degradation.

\begin{table*}[t]
% \scriptsize
% \setlength{\tabcolsep}{3pt}
% \renewcommand{\arraystretch}{1.2}
  \centering
 
  \begin{tabular}{l c ccc ccc ccc r r r}
    \toprule
    \multirow{2}{*}{Method}
    &\multirow{2}{*}{Venue}
    & \multicolumn{3}{c}{\textbf{NUAA-SIRST}}
    & \multicolumn{3}{c}{\textbf{NUDT-SIRST}}
    & \multicolumn{3}{c}{\textbf{IRSTD-1k}}
    & \multirow{2}{*}{Params}
    & \multirow{2}{*}{FLOPs}
    & \multirow{2}{*}{FPS}\\
    \cmidrule(lr){3-5} \cmidrule(lr){6-8} \cmidrule(lr){9-11}
     & & P & R & F1 & P & R & F1 & P & R & F1 &  &  &  \\
    \midrule
    % MDvsFA      & ICCV'2019  & 84.5 & 50.7 & 63.4 & 60.8 & 19.2 & 29.2 & 55.0 & 48.3 & 51.4 & - & - & -\\
    ACM         & WACV'2021  & 76.5 & 76.2 & 76.3 & 73.2 & 74.5 & 73.8 & 67.9 & 60.5 & 64.0 & \underline{3.0} & \textbf{24.7} & 29.1\\
    % FC3Ne       & 93.4 & 88.4 & 90.8 & 97.2 & 91.8 & 94.4 & 85.7 & 79.5 & 82.5 & - & - & -\\
    ISNet       & CVPR'2022  & 82.0 & 84.7 & 83.4 & 74.2 & 83.4 & 78.5 & 71.8 & 74.1 & 72.9 & 3.5 & 265.7 & 10.2\\
    % ALCNe       & 84.8 & 78.0 & 81.3 & 86.8 & 77.2 & 81.7 & 84.3 & 65.6 & 73.8 & - & - & -\\
    DNANet      & TIP'2022   & 84.7 & 83.6 & 84.1 & 91.4 & 88.9 & 90.1 & 76.8 & 72.1 & 74.4 & 7.2 & 135.2 & 7.8\\
    UIUNet      & TIP'2022   & 93.4 & 88.4 & 90.8 & 97.2 & 91.8 & 94.4 & 85.7 & 79.5 & 82.5 & 53.1 & 456.7 & 3.6\\
    IRSAM       & ECCV'2024  & 89.4 & 88.7 & 89.1 & 95.1 & 93.4 & 94.2 & 79.6 &
    \underline{80.2} & 79.9& 12.3 & 76.1 & 7.4 \\
    AGPCNet     & TAES'2023  & 39.0 & 81.0 & 52.7 & 36.8 & 68.4 & 47.9 & 41.5 & 47.0 & 44.1 & 14.9 & 366.2 & 4.8\\
    % EFLNet      & TGRS'2024  & 88.2 & 85.8 & 87.0 & 96.3 & 93.1 & 94.7 & 87.0 & 81.7 & 84.3 & - & - & -\\
    PConv       & AAAI'2025  & \textbf{97.1} & 89.0 & \underline{92.9} & \underline{98.0} & \underline{94.7} & \underline{96.4} & \underline{86.7} & \textbf{80.9} & \underline{83.7} & 23.8 & 88.6 & 58.8\\
    TY-RIST     & ICCV'2025  & 92.9 & \underline{92.1} & 92.5 & 96.8 & \textbf{95.8} & 96.3 & 81.0 & 75.2 & 78.0 & \textbf{2.03} & \underline{37.4} & \textbf{123.1}\\
    SPIRIT      & --         & \underline{94.6} & \textbf{93.8} & \textbf{94.2} & \textbf{98.3} & \underline{94.7} & \textbf{96.5} & \textbf{89.3} & \textbf{80.9} & \textbf{84.9} & 32.2 & 63.7 & \underline{60.1}\\
    \bottomrule
  \end{tabular}
   \caption{Comparison on single-frame benchmarks (NUAA-SIRST, NUDT-SIRST and IRSTD-1k).  The best results are highlighted in bold and the second-place results are highlighted in underline.}
  \label{tab:single_comparison}
\end{table*}

\section{Experiments}

\subsection{Datasets and Evaluation Metrics}
\paragraph{Datasets.} For our experiments, we evaluate the performance of our method using both single-frame and multi-frame datasets. The single-frame datasets used are  NUAA-SIRST~\cite{dai2021asymmetric}, IRSTD-1k~\cite{zhang2022isnet}, and NUDT-SIRST~\cite{li2022dense}. For multi-frame tracking, we use the 
% IRSTD-UAV~\cite{2026AAAI_TDCNet}, 
IRDST~\cite{RDIAN} and IRSTD-15k~\cite{duan2024triple} datasets.
\paragraph{Evaluation Metrics}
For evaluation, we adopt standard metrics, including precision (P), recall (R), F1-score (F1), and average precision (AP50), all computed at an intersection-over-union (IoU) threshold of 0.5. 
% Computational complexity is assessed using the number of parameters (Params) and floating point of operations (FLOPs).
Real-time performance is measured in frames per second (FPS), while computational complexity is assessed using the number of parameters (Params) and floating point of operations (FLOPs).
\subsection{Implementation Details}
The algorithm is implemented in PyTorch and utilizes the AdamW optimizer with $\beta$ values set to [0.9, 0.999]. For parameters related to the dinov3 backbone network, the learning rate is set to $1 \times 10^{-5}$ and weight decay is set to $1 \times 10^{-4}$. For other parts of the model, the optimizer's learning rate is set to $5 \times 10^{-4}$and weight decay is set to 0.000125. For all experiments, the input frames are resized to $512 \times 512$ pixels. Unless stated otherwise, we use DINOv3 ViT-S for the backbone and K=5 for the memory length in multi-frame settings.
% For multi-frame settings, we use video clips of length $T=5$. For single-frame settings, we set $T=1$, reducing the model to per-image inference without temporal aggregation. 
The batch size is 24, and the training runs for 50 epochs.
To supervise the training, we employ bbox loss (L1), GIoU loss \cite{rezatofighi2019generalized}, and cls Loss.
The total loss is given by:
\(
L = \lambda_1 \cdot L_{\text{bbox}} + \lambda_2 \cdot L_{\text{cls}} + \lambda_3 \cdot L_{\text{GIoU}}
\)
where the ratio of the coefficients is set to 2:1:1.
Training is performed using two RTX 3090 GPUs, and testing is conducted with a single RTX 3090 GPU.
For single-frame methods, we include 
% MDvsFA \cite{Wang_2019_ICCV}, 
ACM \cite{dai2021asymmetric}, 
ISNet \cite{zhang2022isnet}, 
DNANet \cite{li2022dense}, 
UIUNet \cite{wu2022uiuNet},      
AGPCNet \cite{zhang2023attention},
% MSHNet   \cite{liu2024MSH}, 
% SCTransNet  \cite{yuan2024sctransnet},
% EFLNet \cite{yang2024eflnet},     
PConv \cite{yang2025pinwheel}    
and TY-RIST \cite{atrash2025ty}. For  multi-frame methods, we include 
DTUM       \cite{li2023direction}, 
TMP    \cite{zhu2024tmp},        
SSTNet   \cite{chen2024sstnet},    
STMENet   \cite{peng2025moving},   
MOCID     \cite{zhang2025mocid}   
and TDCNet   \cite{fang2025spatio}

\begin{figure*}[t]
  \centering
  \includegraphics[width=0.95\linewidth]{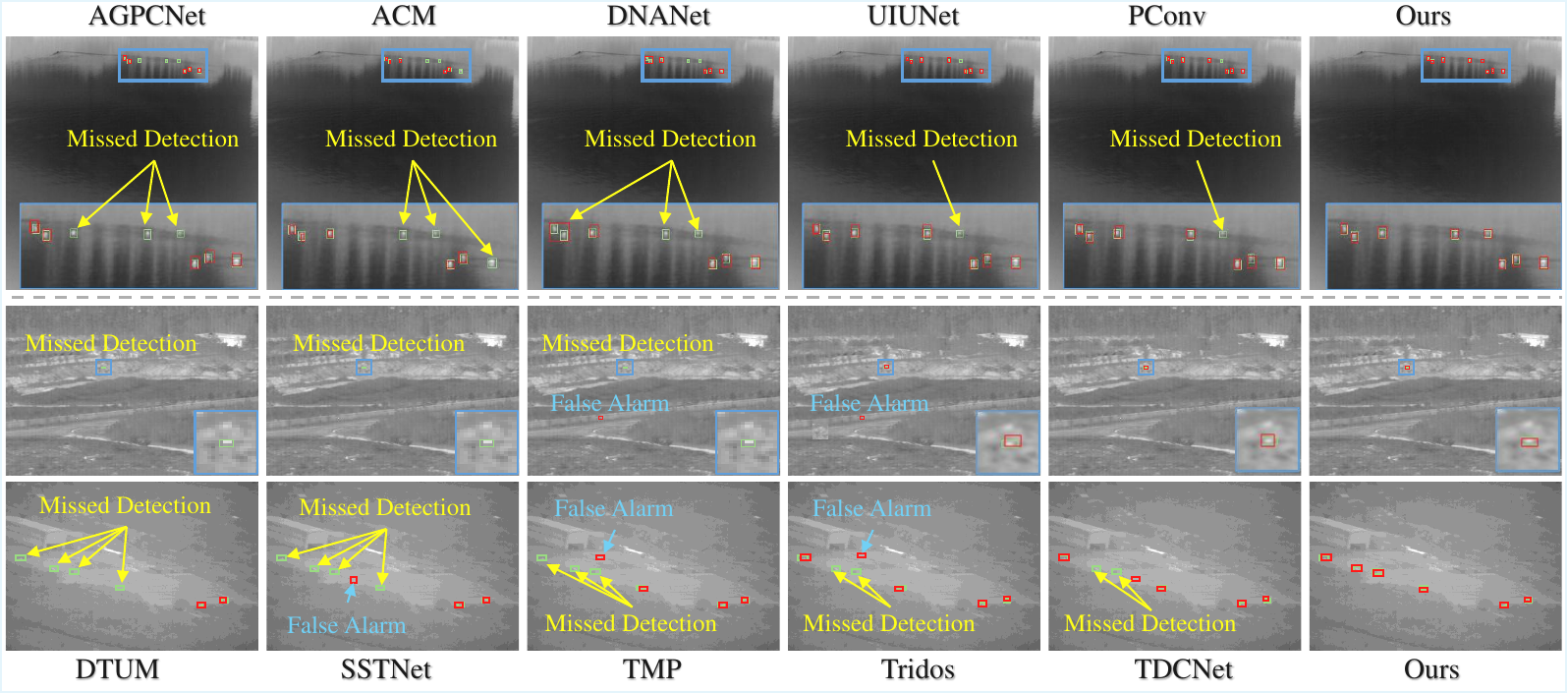}
  \caption{Visual comparison of results from different methods on the IRSTD-1k, IRDST, and IRSTD-15k datasets. Boxes in green and red represent ground-truth and detected targets, respectively.}
  \label{fig:Qualitative}
\end{figure*}
\subsection{Main Results}
\textbf{Multi-frame benchmarks.}
Table~\ref{tab:multi-frame} reports results on two multi-frame benchmarks, IRDST and IRSTD-15k. SPIRIT consistently outperforms previous methods across all metrics. Compared with the strongest prior competitor, SPIRIT achieves notable improvements in both AP50 and F1-score on IRDST. On IRSTD-15k, our method further secures a clear performance lead, demonstrating higher precision and recall compared to existing approaches. In videos, SPIRIT benefits from the joint effect of two complementary components: PIFR enhances sparse target evidence while suppressing globally consistent clutter in the feature domain, and PGMA converts under-constrained appearance matching into feasibility-constrained memory retrieval with confidence-gated updates. Their combination yields more stable detections under infrared appearance ambiguity, reducing spurious associations and preventing error reinforcement over time.

\noindent\textbf{Single-frame benchmarks.}
Table~\ref{tab:single_comparison} compares SPIRIT with representative CNN-based IRSTD detectors on IRSTD-1k, NUAA-SIRST, and NUDT-SIRST. SPIRIT achieves state-of-the-art performance across the evaluated metrics, indicating that the proposed physics-informed feature refinement effectively alleviates feature submergence of tiny, low-SCR targets under structured clutter. Notably, the consistently high F1 scores suggest that our method maintains a favorable balance between precision and recall, which is crucial in IRSTD where both clutter-induced false alarms and missed detections are common.

\noindent\textbf{Qualitative Results}
Figure~\ref{fig:Qualitative} provides qualitative comparisons on representative single-frame and multi-frame IRSTD scenes. Overall, competing methods exhibit typical failure modes in infrared imagery, including missed detections under low-SCR targets, clutter-induced false alarms, and imprecise localization. In contrast, SPIRIT yields more complete detections with fewer spurious responses and tighter boxes, consistent with the quantitative results.
In the single-frame case with multiple targets, several baselines miss at least one target, whereas SPIRIT detects all targets. In the multi-frame single-target case, baselines either miss the target, trigger false alarms, or show less accurate boxes, while SPIRIT maintains stable detection and tighter localization. In the multi-frame multi-target case, baselines suffer from misses or false alarms and degraded localization under multiple distractors, whereas SPIRIT detects all targets with fewer false positives and more precise boxes.

\subsection{Ablation Study}
\textbf{Impact of the proposed PIFR and PGMA.} Table~\ref{tab:ablation} validates the contribution of each component. The Baseline suffers from feature submergence, resulting in suboptimal Recall. The integration of PIFR yields a substantial improvement in Recall, validating that the physics-informed shrinkage effectively recovers submerged sparse targets from deep features. Conversely, PGMA primarily enhances Precision by leveraging temporal feasibility priors to filter out physically implausible dynamic clutter. The best performance is achieved when both modules work in synergy. Furthermore, our proposed framework significantly outperforms the standard Adapter. While the Adapter provides marginal gains through generic domain adaptation, it relies on implicit feature remapping and lacks specific inductive biases for small targets. In contrast, our approach explicitly models the physical characteristics of background low-rankness and temporal feasibility, proving more effective than generic adapters in addressing the fundamental challenges of feature submergence and appearance ambiguity.
\begin{table}[t]
  \centering

  \begin{tabular}{l cccc}
    \toprule
    Setting & AP50 & P & R & F1 \\
    \midrule
    Baseline                  & 86.21 & 91.44 & 89.37 & 90.39 \\
    + Adapter                 & 86.85 & 92.15 & 90.82 & 91.48 \\
    + PIFR                   & \underline{88.62} & 93.28 & \underline{96.55} & \underline{94.89} \\
    + PGMA                    & 87.14 & 94.62 & 89.95 & 92.23 \\
    + Adapter + PGMA          & 87.90 & \underline{95.10} & 91.35 & 93.19 \\
    + PIFR + PGMA            & \textbf{89.56} & \textbf{95.34} & \textbf{97.15} & \textbf{96.23} \\
    \bottomrule
  \end{tabular}
\caption{Ablation study on {IRSTD-15k}. ''Adapter'' denotes the standard parameter-efficient fine-tuning module proposed by ~\cite{houlsby2019parameter}.}
  \label{tab:ablation}
\end{table}
\begin{figure}[t]
  \centering
  \includegraphics[width=0.95\linewidth]{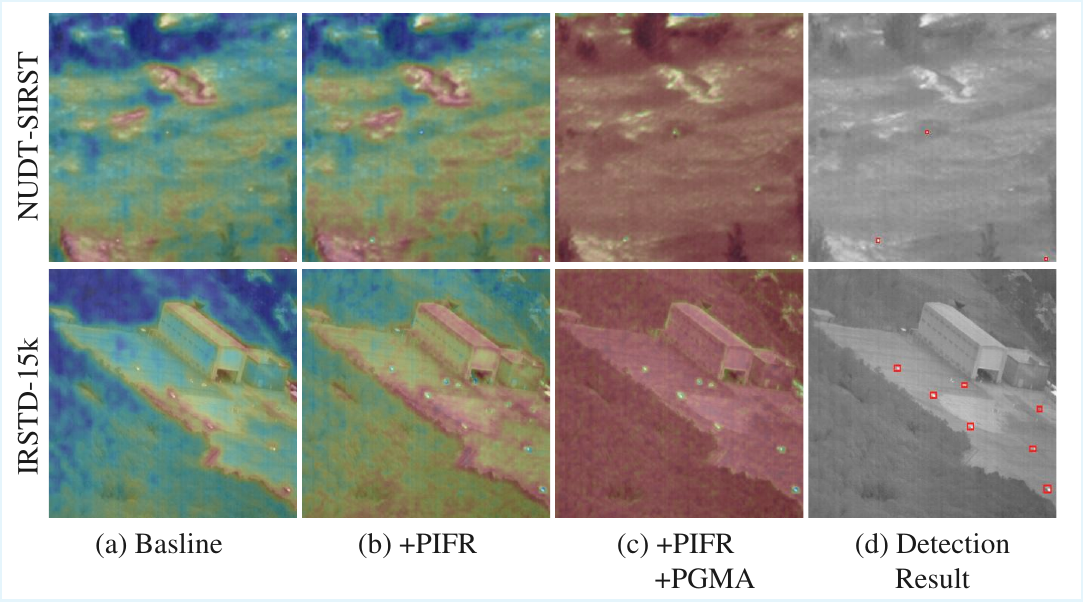}
  \caption{Heat map visualization illustrating the progressive enhancement achieved by PIFR and PGMA.}
  \label{fig:ablation}
\end{figure}

Figures~\ref{fig:ablation} show feature visualizations for single-frame and multi-frame scenarios. In the single-frame case, applying {PIFR} clearly enhances target features, making them more salient and less affected by clutter. PGMA has little effect on single-frame detection, keeping performance stable without temporal context.
For multi-frame sequences, {PGMA} helps filter spurious targets and reduce false positives, as reflected in the visualizations. Target features become clearer, and the model better separates real targets from background clutter. This indicates {PGMA} effectively stabilizes temporal associations and suppresses tracking errors, resulting in more robust and accurate video detection.

\begin{figure}[t]
  \centering
  \includegraphics[width=0.85\linewidth]{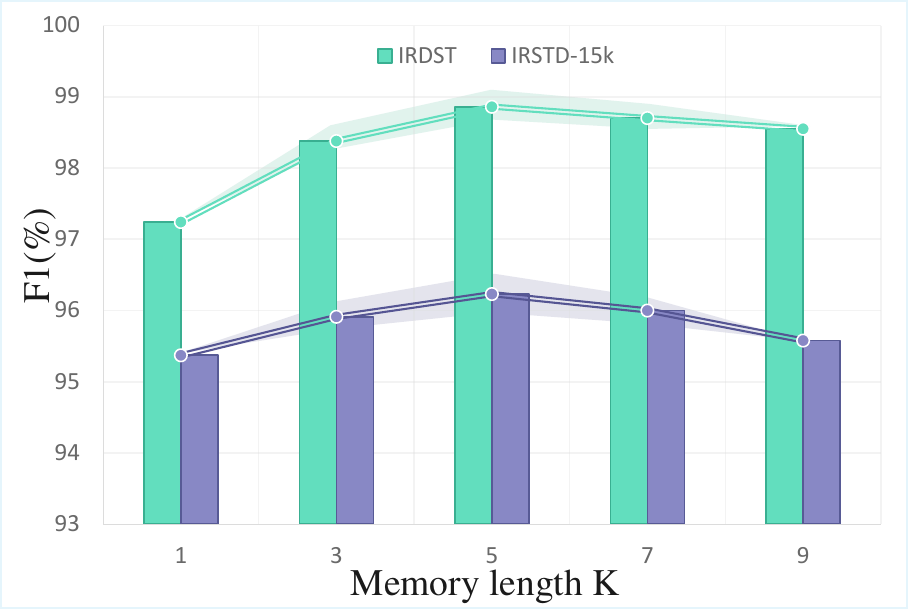}
  \caption{Sensitivity to memory length $K$ on multi-frame datasets.}

  \label{fig:K_sensitivity}
\end{figure}

\textbf{Impact of the memory length K.} 
We investigate the sensitivity of SPIRIT to the memory length $K$ by varying  $K \in \{1,3,5,7,9\}$ while keeping the input resolution fixed.
Figure~\ref{fig:K_sensitivity} shows that increasing $K$ from 1 to a moderate value consistently improves performance, indicating that retaining short-term history benefits feasibility-constrained retrieval under infrared appearance ambiguity. However, when $K$ becomes too large, the performance saturates and can even drop. This trend is consistent with PGMA: moderate memory provides relevant recent evidence to recover weak targets across frames, whereas an overly long memory introduces a larger pool of older key-value tokens that are less relevant to the current frame. Since infrared backgrounds are highly non-stationary, stale memories may contain outdated clutter patterns that remain visually target-like and act as strong distractors during retrieval. We therefore set $K=5$ by default as a good trade-off between accuracy and efficiency.

\begin{figure}[t]
  \centering
  \includegraphics[width=0.9\linewidth]{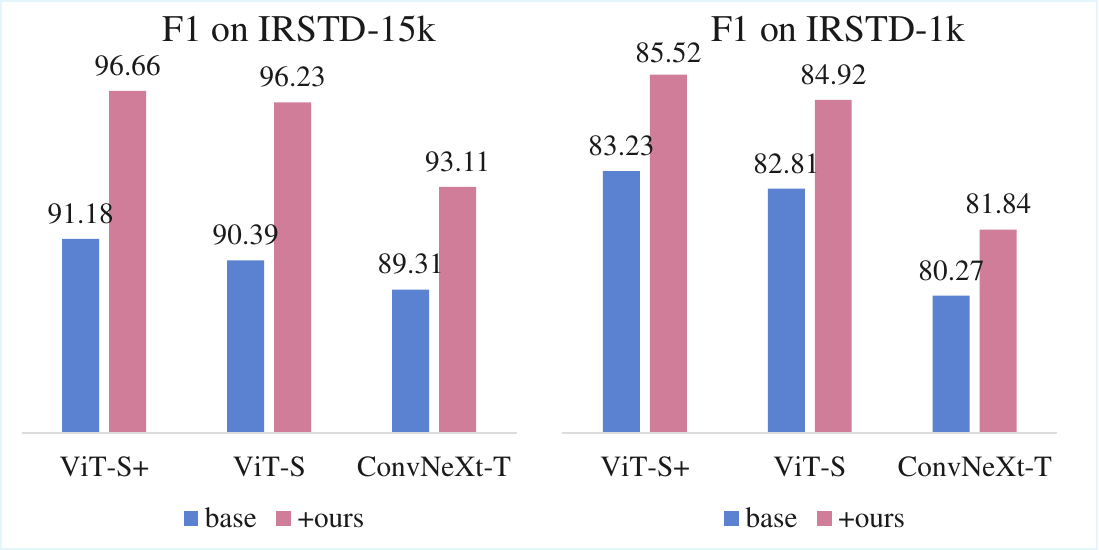}
  \caption{Performance comparison across different backbones.}

  \label{fig:backbone}
\end{figure}
\paragraph{Generalization across Backbones.}
To verify the universality of our framework, we evaluate the proposed method with three different backbones: DINOv3-based  ViT-S+,  ViT-S, and ConvNeXt-T, on both IRSTD-15k and IRSTD-1k datasets. As shown in Figure~\ref{fig:backbone}, performance naturally improves with stronger backbones. Integrating our proposed modules yields consistent and significant improvements across all architectures. This demonstrates that our method is not reliant on a specific backbone but addresses the limitations of deep features in IRSTD, providing a robust plug-and-play solution for diverse foundation models.
\section{Conclusion}
In this work, we propose SPIRIT, a unified framework that adapts Vision Foundation Models for robust infrared small target detection. The proposed approach overcomes the limitations of generic semantic features through two lightweight, physics-inspired plug-ins. Specifically, PIFR mitigates feature submergence by approximating a rank-sparsity decomposition to enhance target discriminability, while PGMA resolves appearance ambiguity by injecting spatial feasibility priors into memory attention. This design enables a single model to handle both video and single-frame inputs effectively. Experimental results across multiple benchmarks confirm that SPIRIT significantly outperforms existing baselines and state-of-the-art methods, validating the effectiveness of embedding physical constraints into large-scale pretrained backbones.
% \clearpage
\bibliographystyle{named}
\bibliography{ijcai26}
\end{CJK}
\end{document}